\documentclass[acmtog]{acmart}

\acmSubmissionID{1094}

\usepackage[ruled]{algorithm2e} 

\SetAlFnt{\small}
\SetAlCapFnt{\small}
\SetAlCapNameFnt{\small}
\SetAlCapHSkip{0pt}

\usepackage{bm} 
\usepackage{subcaption} 
\usepackage[nomessages]{fp}

\copyrightyear{2025}
\acmYear{2025}
\setcopyright{rightsretained}
\acmConference[SA Conference Papers '25]{SIGGRAPH Asia 2025 Conference Papers}{December 15--18, 2025}{Hong Kong, Hong Kong}
\acmBooktitle{SIGGRAPH Asia 2025 Conference Papers (SA Conference Papers '25), December 15--18, 2025, Hong Kong, Hong Kong}\acmDOI{10.1145/3757377.3763829}
\acmISBN{979-8-4007-2137-3/2025/12}

\begin{document}
\title{TrackerSplat: Exploiting Point Tracking for Fast and Robust Dynamic 3D Gaussians Reconstruction}

\author{Daheng Yin}
\orcid{0000-0002-9431-4240}
\affiliation{%
 \institution{Simon Fraser University}
 \city{Burnaby}
 \country{Canada}}
\email{dya64@sfu.ca}

\author{Isaac Ding}
\orcid{0009-0004-1117-3860}
\affiliation{%
 \institution{Simon Fraser University}
 \city{Burnaby}
 \country{Canada}}
\email{isaac_ding@sfu.ca}

\author{Yili Jin}
\orcid{0000-0002-7127-8902}
\affiliation{%
\institution{McGill University}
\city{Montreal}
\country{Canada}}
\affiliation{%
 \institution{Simon Fraser University}
 \city{Burnaby}
 \country{Canada}}
\email{yili.jin@mail.mcgill.ca}

\author{Jianxin Shi}
\orcid{0000-0002-7687-8480}
\affiliation{%
\institution{Nankai University}
\city{Tianjin}
\country{China}}
\affiliation{%
 \institution{Simon Fraser University}
 \city{Burnaby}
 \country{Canada}}
\email{jxshi@nankai.edu.cn}

\author{Jiangchuan Liu}
\orcid{0000-0001-6592-1984}
\affiliation{%
\institution{Simon Fraser University}
\city{Burnaby}
\country{Canada}}
\email{jcliu@cs.sfu.ca}

\begin{abstract}

Recent advancements in 3D Gaussian Splatting (3DGS) have demonstrated its potential for efficient and photorealistic 3D reconstructions, which is crucial for diverse applications such as robotics and immersive media.
However, current Gaussian-based methods for dynamic scene reconstruction struggle with large inter-frame displacements, leading to artifacts and temporal inconsistencies under fast object motions.
To address this, we introduce \textit{TrackerSplat}, a novel method that integrates advanced point tracking methods to enhance the robustness and scalability of 3DGS for dynamic scene reconstruction.
TrackerSplat utilizes off-the-shelf point tracking models to extract pixel trajectories and triangulate per-view pixel trajectories onto 3D Gaussians to guide the relocation, rotation, and scaling of Gaussians before training.
This strategy effectively handles large displacements between frames, dramatically reducing the fading and recoloring artifacts prevalent in prior methods.
By accurately positioning Gaussians prior to gradient-based optimization, TrackerSplat overcomes the quality degradation associated with large frame gaps when processing multiple adjacent frames in parallel across multiple devices, thereby boosting reconstruction throughput while preserving rendering quality.
Experiments on real-world datasets confirm the robustness of TrackerSplat in challenging scenarios with significant displacements, achieving superior throughput under parallel settings and maintaining visual quality compared to baselines.
The code is available at \href{https://github.com/yindaheng98/TrackerSplat}{https://github.com/yindaheng98/TrackerSplat}.

\end{abstract}

\ccsdesc[500]{Computing methodologies~Rendering}
\ccsdesc[500]{Computing methodologies~Image-based rendering}
\ccsdesc[300]{Computing methodologies~Point-based models}
\ccsdesc[300]{Computing methodologies~Motion capture}

\keywords{Point Tracking, 3D Gaussian Splatting}

\begin{teaserfigure}
	\centering
	\begin{subfigure}[t]{0.24\linewidth}
        \includegraphics[width=\linewidth,trim={195mm 0 78mm 0},clip]{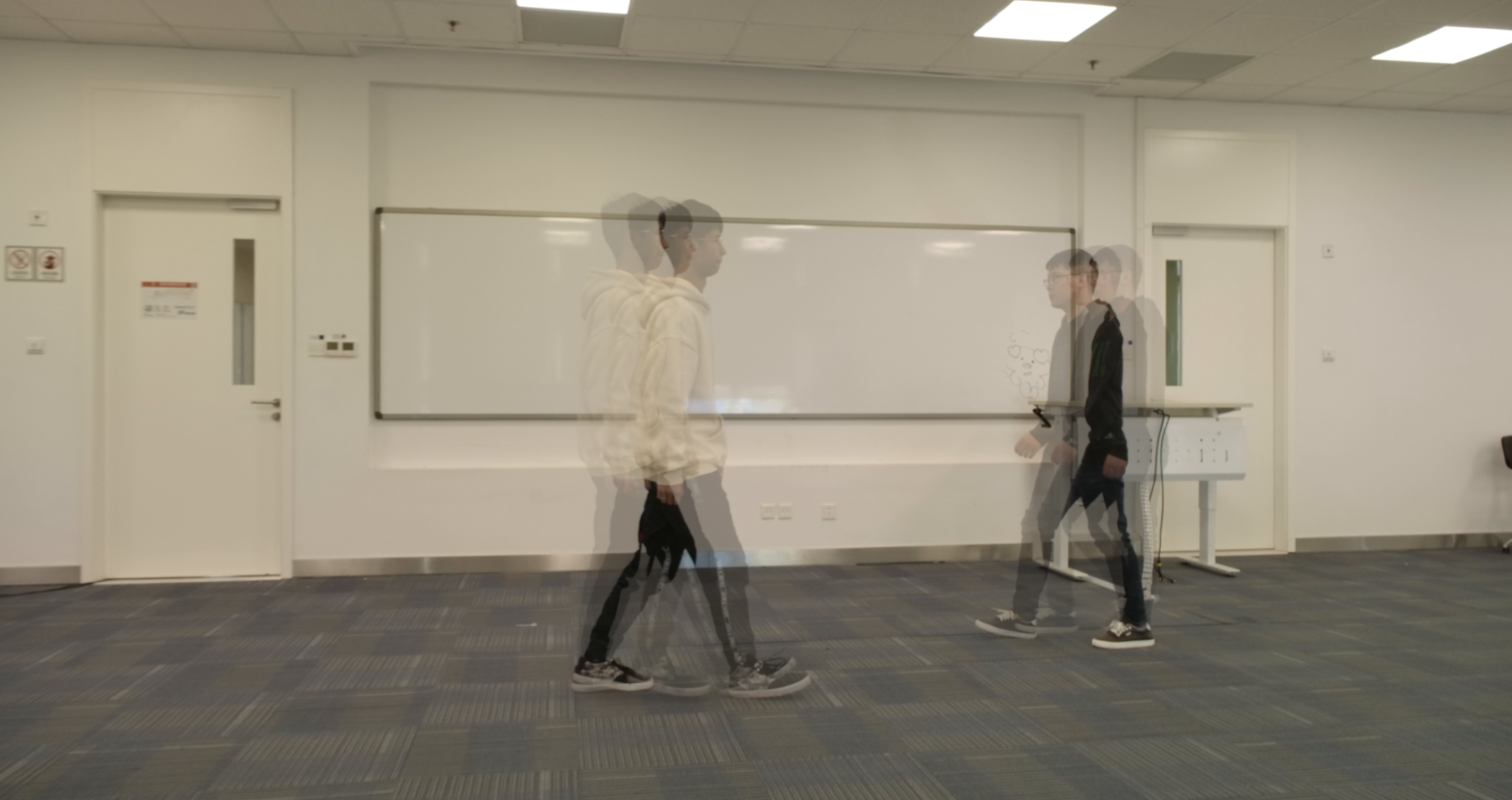}
        \caption{Ground truth frame 61, 64, 67.}\label{fig:fade1}
    \end{subfigure}
	\begin{subfigure}[t]{0.24\linewidth}
        \includegraphics[width=\linewidth,trim={25mm 0 10mm 0},clip]{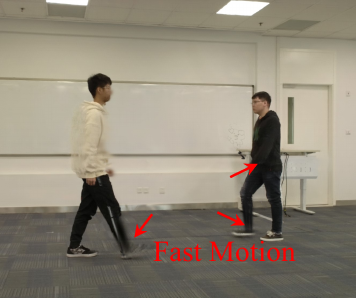}
        \caption{Frame 64 by Dynamic3DGS.}\label{fig:fade3}
    \end{subfigure}
	\begin{subfigure}[t]{0.24\linewidth}
        \includegraphics[width=\linewidth,trim={25mm 0 10mm 0},clip]{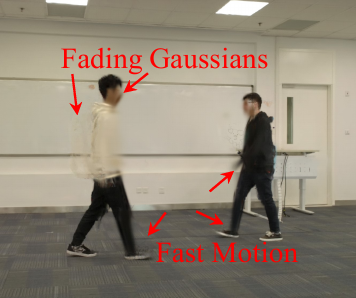}
        \caption{Frame 67 by Dynamic3DGS.}\label{fig:fade5}
    \end{subfigure}
	\begin{subfigure}[t]{0.24\linewidth}
        \includegraphics[width=\linewidth,trim={195mm 0 78mm 0},clip]{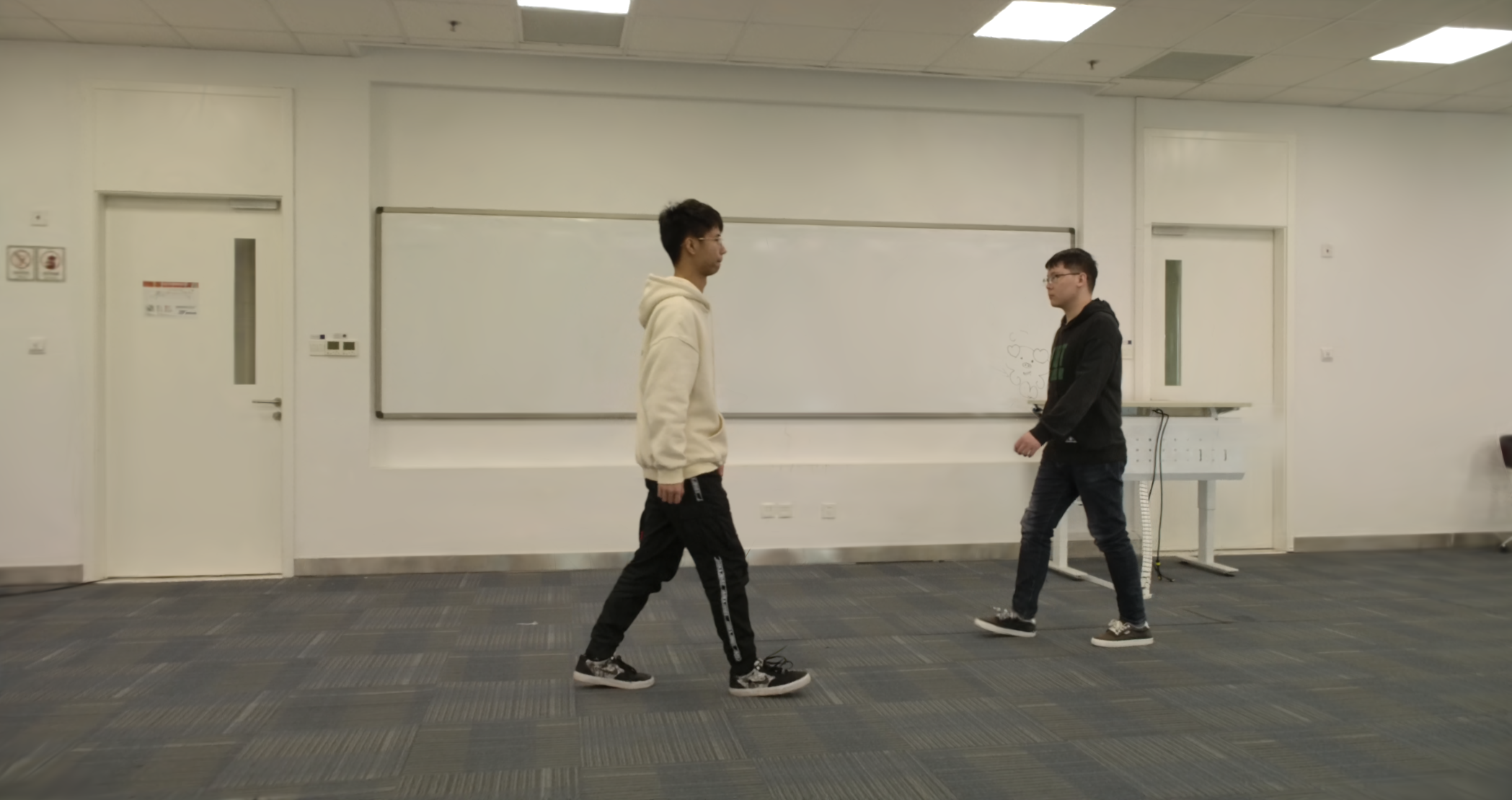}
        \caption{Frame 67 by our method.}\label{fig:ours}
    \end{subfigure}\\
	\begin{subfigure}[t]{.29\linewidth}
        \includegraphics[width=\linewidth,trim={0 4.8mm 0 0},clip]{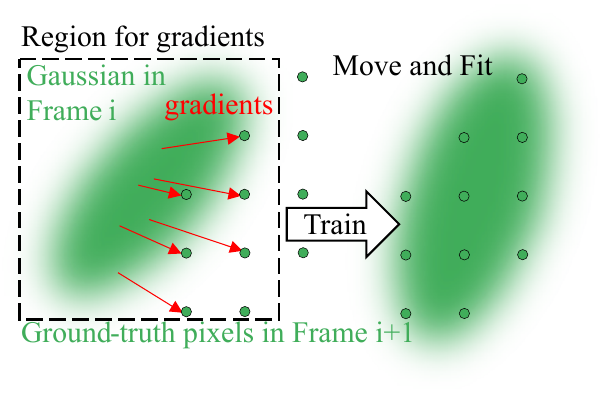}
		\vspace*{-6mm}
        \caption{Gaussian fits for slow motion.}\label{fig:fit}
    \end{subfigure}
	\begin{subfigure}[t]{.29\linewidth}
        \includegraphics[width=\linewidth]{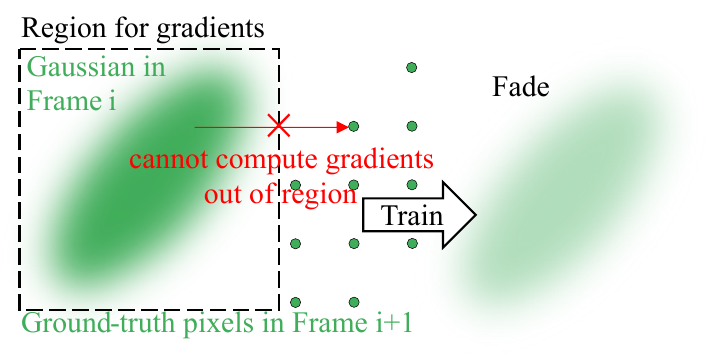}
		\vspace*{-6mm}
        \caption{Gaussian fades for fast motion.}\label{fig:fade}
    \end{subfigure}
	\begin{subfigure}[t]{.40\linewidth}
        \includegraphics[width=\linewidth,trim={0 7mm 0 0},clip]{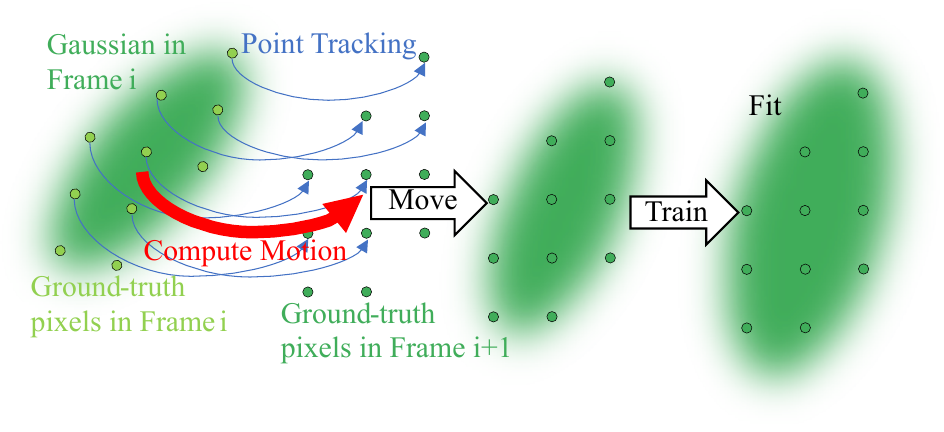}
		\vspace*{-6mm}
        \caption{Basic idea of TrackerSplat}\label{fig:fit-fade-ours}
    \end{subfigure}
    \caption{
        Illustration of the motivation and basic idea of TrackerSplat.
        (a) Ground truth from the "walking" sequence.
        (b), (c) Rendered frames 64 and 67, trained 1000 iterations from frame 61 with the physically-based regularization losses introduced by Dynamic3DGS~\cite{luiten2023dynamic}. Gaussians struggle to accurately follow the fast moving object, resulting in fading or incorrect recoloring.
        (e) Slow motion: object remains in the region for gradient computing, allowing Gaussians to maintain consistent color and follow the movement of object.
        (f) Fast motion: the object moves outside the region for gradient computing, position gradients fail to align with the movement of object, causing Gaussians to either fade or incorrectly recolor.
        (d), (g) TrackerSplat to adjust Gaussian position, rotation and scaling parameters according to point tracking results before training, enabling stable and robust training for fast-moving objects.
    }
\end{teaserfigure}

\maketitle

\section{Introduction}

Reconstructing dynamic 3D scenes and generating photo-realistic, temporally consistent renderings have long been fundamental goals in computer vision and graphics.
These capabilities are increasingly important for creating controllable, editable, high-quality 3D content, underpinning applications in film, gaming, and the metaverse~\cite{zhangEditableFreeviewpointVideo2021}.
Beyond visual fidelity for human audiences, a more critical requirement emerges in embodied AI and robotics: capturing the trajectories of objects and agents in dynamic environments.
For robots, temporally coherent reconstruction provides the motion cues necessary for downstream tasks such as object manipulation, navigation, and human-robot interaction~\cite{abou-chakraPhysicallyEmbodiedGaussian2024}.
It also enables reliable tracking and prediction of surrounding agents in applications like autonomous driving~\cite{zhouDrivingGaussianCompositeGaussian2024}.
Despite these increasing demands, efficient and accurate reconstruction of dynamic scenes while preserving consistent object trajectories remains challenging due to the complexities introduced by temporal dynamics and diverse motion patterns.

Recent progress in 3D reconstruction has been driven notably by the success of 3D Gaussian Splatting (3DGS)\cite{kerbl3DGaussianSplatting2023} for its ability to efficiently represent 3D scenes with photorealism.
By modeling 3D space with ellipsoids (``Gaussians''), 3DGS enables intuitive editing through manipulation of individual Gaussians, making it suitable for dynamic scene representation.
Building on its strengths, follow-up studies has adapted 3DGS to dynamic scenes by explicitly encoding Gaussian parameters as temporal trajectories~\cite{liSpacetimeGaussianFeature2024,linGaussianFlow4DReconstruction2024} or by representing motion fields using implicit features~\cite{wu4DGaussianSplatting2024,liST4DGSSpatialTemporallyConsistent2024}.
These methods typically rely on frame-to-frame adaptation, iteratively refining Gaussian parameters by training on consecutive frames to ensure smooth temporal transitions~\cite{gaoHiCoMHierarchicalCoherent2024,luiten2023dynamic}.

Despite these advancements, the reconstruction process of 3DGS is computationally intensive, limiting its application in scenarios demanding both high quality and high throughput, such as live streaming and interactive virtual environments.
To improve the throughput of reconstruction without adding end-to-end latency, a natural solution is to process multiple adjacent frames in parallel across multiple GPUs.
However, our experiments reveal that existing methods suffer from significant quality degradation when handling large displacements between frames, leading to visible artifacts, as is shown in Figure~\ref{fig:fade3} and Figure~\ref{fig:fade5}.

Upon further analysis, we identify a critical issue in 3DGS that contributes to this quality degradation.
Existing approaches rely heavily on fine-tuning Gaussian parameters from frame to frame using iterative training.
A core idea behind these methods is to guide Gaussian motion using position gradients computed by comparing Gaussian colors with the surrounding pixels they overlap (Figure~\ref{fig:fit}).
Due to computational constraints, gradient computations are restricted to a limited local neighborhood.
This constraint results in inaccurate position gradients when objects experience significant inter-frame motion and move outside this restricted region (Figure~\ref{fig:fade}).
In parallel setups, increased parallelism widens the frame gaps assigned to each device, amplifying the likelihood of significant object displacements and consequently exacerbating this issue, leading to prominent artifacts (Figure~\ref{fig:fade5}).

To mitigate this limitation, we propose directly estimating Gaussian trajectories across frames rather than relying solely on the gradient to update their positions.
Recent advancements in point tracking~\cite{karaevCoTrackerItBetter2025,karaevCoTracker3SimplerBetter2024} provide robust pixel-level motion estimation across video frames.
However, integrating point tracking presents two key challenges: (1) translating 2D pixel trajectories into updates for 3D Gaussian parameters, and (2) mitigating inaccuracies in pixel trajectories to prevent error accumulation during updates.

We introduce \textit{TrackerSplat} to address these challenges.
As illustrated in Figure~\ref{fig:fit-fade-ours}, TrackerSplat integrates an off-the-shelf point tracking model to capture pixel trajectories for each viewpoint.
To compute updates for 3D Gaussians, we propose Parallel Weighted Incremental Least Squares (PWI-LS) that derives 2D motion from pixel trajectories.
These 2D motions from multiple views are triangulated to update Gaussian positions, rotations, and scales.
To reduce inaccuracies, the computed updates are smoothed by Motion Regularization, and Gaussian parameters are further refined through training.
By repositioning Gaussians closer to their correct locations before training, TrackerSplat maintains coherent tracking despite large displacements, significantly reducing fading or recoloring artifacts observed in prior methods.
Most importantly, since tracking mitigates the impact of large frame gaps, TrackerSplat enables independent frame updates across multiple GPUs, thus increasing reconstruction throughput without sacrificing quality.

To the best of our knowledge, we are the first to identify the robustness limitations of 3DGS in handling large inter-frame displacements, and TrackerSplat is the first method to directly compute Gaussian trajectories using multi-view point tracking results to address this limitation.
While prior works have incorporated tracking within Gaussian-based pipelines~\cite{stearnsDynamicGaussianMarbles2024,leiMoScaDynamicGaussian2025}, direct use of multi-view point tracking for trajectory estimation remains unexplored.

We implement TrackerSplat with a parallel pipeline across 8 GPUs and evaluate it on real-world dynamic scene datasets.
Our extensive experiments demonstrate superior throughput under parallel settings, while preserving or improving visual quality compared to baselines.
Our findings confirm the effectiveness of incorporating point tracking into 3DGS-based dynamic reconstruction, paving the way for scalable, accurate, and temporally consistent dynamic 3D scene reconstructions.

\section{Related Work}

\subsection{3D Gaussian Splatting for Dynamic Scenes}

Recent years have witnessed significant progress in reconstructing 3D representations from multi-view captures.
Among these advancements, 3D Gaussian Splatting (3DGS)~\cite{kerbl3DGaussianSplatting2023} has emerged as a leading approach.
3DGS represents scenes using a set of ellipsoids ("Gaussians") and achieves photorealistic rendering with high efficiency.
Building on its success, the latest studies have extended 3DGS to dynamic scenarios.
Some methods dynamically add or remove Gaussians to represent motion through their appearance and disappearance~\cite{duan4DRotorGaussianSplatting2024,sun3DGStreamFlyTraining2024}.
However, this approach may lead to significant storage overhead due to the large number of Gaussians needed to capture the dynamic nature of the scene.
To address this, other methods explicitly represent dynamic scenes using Gaussians and their trajectories, significantly reducing the number of Gaussians.
For instance, some approaches use implicit features conditioned on time to represent motion fields~\cite{luiten2023dynamic,wu4DGaussianSplatting2024,liST4DGSSpatialTemporallyConsistent2024,liSpacetimeGaussianFeature2024,linGaussianFlow4DReconstruction2024,gaoHiCoMHierarchicalCoherent2024}, while others adopt triplane representations for higher spatial and temporal resolutions~\cite{wu4DGaussianSplatting2024}.
Other advancements also include frame-to-frame adaptation techniques, where Gaussian parameters are iteratively refined by training on consecutive frames~\cite{gaoHiCoMHierarchicalCoherent2024,luiten2023dynamic,xuRepresentingLongVolumetric2024}.

\subsection{Point Tracking}
\begin{figure}[htbp]
	\centering
	\includegraphics{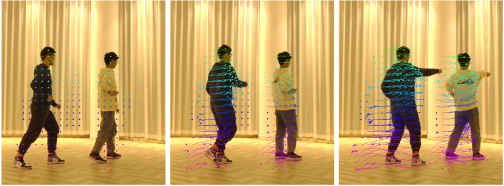}
    \caption{
		DOT point tracking on a video sequence.
		Colored lines show pixel trajectories over time.
	}
    \label{fig:cotracker}
\end{figure}

Point tracking~\cite{wangTrackingEverythingEverywhere2023,seidenschwarzDynOMoOnlinePoint2025} identifies the position and visibility of specific pixels across video sequences (Figure~\ref{fig:cotracker}), providing robust trajectory estimation even under challenging conditions such as occlusion.
Recently, point tracking has attracted considerable attention within the computer vision community.
Empowered by semi-supervised correspondence, CoTracker~\cite{karaevCoTracker3SimplerBetter2024,karaevCoTrackerItBetter2025} achieves state-of-the-art sparse tracking performance, while DOT~\cite{lemoingDenseOpticalTracking2024} further enhances dense tracking accuracy and efficiency.
Point tracking has been used in 3D reconstruction, especially for reconstructing dynamic scenes from monocular videos~\cite{leiMoScaDynamicGaussian2025,stearnsDynamicGaussianMarbles2024}.
In these methods, point tracking typically separates static background from dynamic foreground or serves as a regularization term for dynamic regions.
However, the direct application of point tracker results to compute 3D Gaussian splatting parameters for multi-view dynamic scene reconstruction remains largely unexplored.

\section{Preliminaries}

\subsection{Mathamatical Reperesentation of 3D Gaussians}

3DGS represents the scenes with 3D Gaussians.
Each 3D Gaussians is characterized by two key components: its mean $\bm\mu_{3D}$ represents the position of the ellipsoid, and its covariance matrix $\Sigma_{3D}=RSS^\top R^\top$, composed of a scaling matrix $S$ and a rotation matrix $R$, describes the spread and orientation of the Gaussian ellipsoid respectively.

When projected onto the 2D image plane, the 3D Gaussian becomes a 2D Gaussian distribution.
Concretely, for a point on the image plane, the density function of the 2D Gaussian distribution can be represented as:
\begin{equation}\label{eq:3dgs}
\begin{aligned}
G(\bm x)&=e^{-\frac{1}{2}(\bm x-\bm \mu_{2D})^\top\Sigma_{2D}^{-1}(\bm x-\bm \mu_{2D})}\\
\Sigma_{2D}&=JW\Sigma_{3D} W^\top J^\top\\
\bm \mu_{2D}&=\frac{1}{z}PW\bm\mu_{3D}
\end{aligned}
\end{equation}
where $J$ is the Jacobian of the affine approximation of the projective transformation, $W$ is the viewport transformation matrix, $P$ is the projection transformation matrix, and $z$ is the depth value in $PW\bm\mu_{3D}$.
After projection, the color of each pixel is calculated by alpha-blending each Gaussian according to its depth.

\subsection{Integration of Point Tracking}

To track 3D Gaussians, we rely on point tracking in video frames captured from multiple viewpoints.
In particular, we employ Dense Optical Tracking (DOT)~\cite{lemoingDenseOpticalTracking2024}, a simple yet efficient method for point tracking.
For each pixel $i$ located at position $\bm x_i$ in the first frame, point tracking estimates its corresponding position $(\bm x_i \mapsto \bm x_i')$ in any subsequent target frame.

\section{Method}

\subsection{Overview}

\begin{figure*}[htbp]
	\centering
	\includegraphics{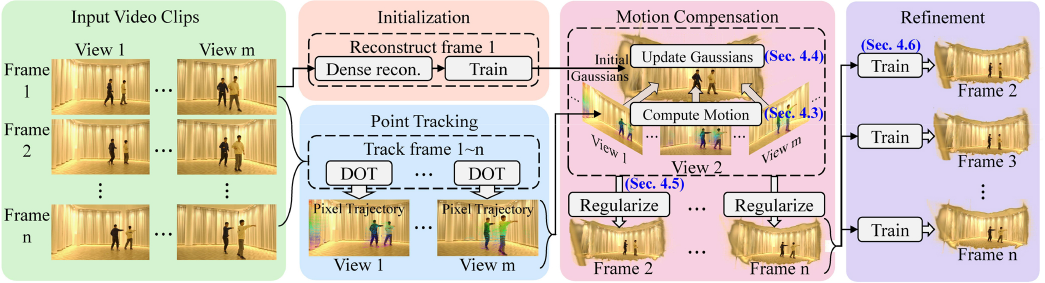}
    \caption{
        TrackerSplat overview.
        Our method processes video clips captured from multiple fixed viewpoints.
        It begins by applying existing reconstruction techniques to initialize a set of 3D Gaussians for the first frame.
        For subsequent frames, the position, rotation, and scale of these Gaussians are updated based on point tracking across views, with their motions regularized by neighboring Gaussians.
        Finally, the Gaussian parameters of each frame are refined by training on input frames.}
    \label{fig:overview}
\end{figure*}

We provide an overview of TrackerSplat in Figure~\ref{fig:overview}.
Our goal is to reconstruct dynamic scenes from video clips captured from multiple fixed viewpoints by tracking and refining a set of Gaussian representations.
Our methods can be divided into four stages:

\paragraph{Initialization}
For the first frame, we initialize 3D Gaussians using existing reconstruction methods for static scenes, such as InstantSplat~\cite{fanInstantSplatUnboundedSparseview2024}.

\paragraph{Point Tracking}
For each subsequent frame, we employ a point tracking model to extract 2D trajectories for every pixel in the initial frame throughout the video clip.

\paragraph{Motion Compensation}
In this stage, we compute the updated parameters of the 3D Gaussians based on the point tracking results.
TrackerSplat first solves the Gaussian motions (Sec.~\ref{sec:gaussian-motion}) on the 2D image plane using \textit{Parallel Weighted Incremental Least Squares (PWI-LS)} (Sec.~\ref{sec:parallel-weighted-incremental-least-squares}) and then \textit{updates Gaussians from multi-view observations} (Sec.\ref{sec:multi-view-motion}).
To mitigate the impact of errors from point tracking and PWI-LS on quality, we introduce \textit{Motion Regularization} (Sec.~\ref{sec:motion-regularization}), which applies median filtering and propagates motion information according to neighboring Gaussians.

\paragraph{Refinement}
Finally, we refine the parameters of each frame by training on the input video clips (Sec.~\ref{sec:refinement}).

In this section, we detail each step of the proposed approach and show how they can be parallelized on multi-GPU devices with our \textit{Parallel Pipeline} (Sec.~\ref{sec:parallel-pipeline}).

\subsection{Definition of Gaussian Motion}\label{sec:gaussian-motion}

To serve as the basis for tracking 3D Gaussians across multiple video viewpoints, it is necessary to define the motion of a 2D Gaussian distribution on the image plane.
Consider a 2D Gaussian distribution $G(\bm x)$ on a specific image plane in the first frame, characterized by its covariance $\Sigma_{2D}$ and mean $\bm{\mu}_{2D}$.
We define the motion of this Gaussian as an affine transformation $[A|\bm b]$, which maps $G(\bm{x})$ to a new 2D Gaussian distribution $G'(\bm{x})$ in a subsequent frame. Formally, $G(\bm x)=G'(A\bm x+\bm b)$.
Under this affine transformation, the updated mean $\bm{\mu}_{2D}'$ and covariance $\Sigma_{2D}'$ can be derived as follows:
\begin{equation}\label{eq:2d-transformation}
\begin{aligned}
\Sigma_{2D}'=A\Sigma_{2D}A^\top\\
\bm\mu_{2D}'=A\bm\mu_{2D}+\bm b
\end{aligned}
\end{equation}

\subsection{Parallel Weighted Incremental Least Squares (PWI-LS)}\label{sec:parallel-weighted-incremental-least-squares}

\subsubsection{Problem Formulation}
Let $\{\bm x_i\},i\in[1,n]$ denote the collection of pixels on a specific image plane covered by the 2D Gaussian in the first frame.
Our goal is to find the affine transformation $[A|\bm{b}]$ that best aligns these pixel coordinates $\bm{x}_i$ with their tracked positions $\bm{x}_i'$ in the subsequent frame.
This can be formulated as follows:
$$\mathop{min}\limits_{A,\bm b}\sum_{i=1}^{n}||A\bm x_i+\bm b-\bm x_i'||^2$$

\subsubsection{Naive Least Squares}
A straightforward method to solve this optimization is through least squares.
By stacking all point pairs $(\bm x_i \mapsto \bm x_i')$ into matrices $X$ and $Y$, the affine transformation $[A|\bm b]$ can be computed as:

\begin{equation}\label{eq:least-squares}
\begin{aligned}
&[\hat A|\hat{\bm b}]=(X^\top X)^{-1}X^\top Y\\
&\begin{aligned}
X&=
\begin{bmatrix}
x_1&\dots&x_i&\dots&x_n\\
1&\dots&1&\dots&1\\
\end{bmatrix}^\top\\
Y&=\begin{bmatrix}x_1'&\dots&x_i'&\dots&x_n'\end{bmatrix}^\top\\
\end{aligned}
\end{aligned}
\end{equation}

However, in cases where many pixels fall under each Gaussian and when multiple Gaussians overlap, explicitly constructing and inverting these large matrices per Gaussian is computationally expensive.
This motivates a more efficient, incremental approach.

\subsubsection{Incremental Least Squares}
To mitigate high computational costs, we exploit the additive structure of $X^\top X$ and $X^\top Y$ in Equation~\ref{eq:least-squares} and decompose them into per-pixel contributions:

\begin{alignat*}{2}
 X^\top X&=\sum_{i=1}^nP_i&&=\sum_{i=1}^n\begin{bmatrix}x_i\\1\end{bmatrix}\cdot[x_i^\top, 1]\\
 X^\top Y&=\sum_{i=1}^nQ_i&&=\sum_{i=1}^n\begin{bmatrix}x_i\\1\end{bmatrix}\cdot x_i'^\top
\end{alignat*}

By expressing the solution as sums of $P_i$ and $Q_i$, the contribution of each pixel can be handled independently.
This incremental scheme avoids building full matrices for all pixels, allowing us to accumulate partial results $P_i$ and $Q_i$ in parallel, paving the way for efficient parallel implementations.

\subsubsection{Weighted Least Squares}
In regions of partial coverage (e.g., object boundaries with overlapping foreground/background Gaussians), unweighted alignment can be misled by pixels that do not truly belong to a particular motion.
For example, consider a static background Gaussian and a high-density moving Gaussian in front of it.
The moving Gaussian may partially cover a moving pixel with very high opacity, and a static pixel with very low opacity.
In this situation, the static pixel should not contribute to the motion of the moving Gaussian.
To achieve this, we adopt a weighted formulation:
\begin{equation}\label{eq:weighted-least-squares}
\begin{gathered}
 \relax[\hat A|\hat{\bm b}]=V_1^{-1}V_2\\
 V_1=\sum_{i=1}^nw_iP_i \quad V_2=\sum_{i=1}^nw_iQ_i\\
\end{gathered}
\end{equation}
where each pixel carries a weight $w_i$ representing the likelihood that its motion corresponds to a particular Gaussian.
In our implementation, $w_i$ is set to $\alpha_i T_i$, where $\alpha_i$ and $T_i$ are the opacity and the transparency of Gaussian at the pixel $i$ in alpha-blending.
This ensures that boundary pixels with low opacity contribute less to the accumulated statistics, thereby reducing sensitivity to irrelevant motions from overlapping regions.

\subsubsection{Acceleration by GPU}

The Weighted Incremental Least Squares algorithm can be optimized for GPU execution to exploit its parallel processing capabilities for acceleration, which results in the \textit{Parallel Weighted Incremental Least Squares} algorithm:

For each Gaussian, we allocate GPU memory for the $(3\times 3)$ matrix $V_1$ and the $(3\times 2)$ matrix $V_2$.
We modify the rendering process to compute $V_1$ and $V_2$, where the $w_iP_i$ and $w_iQ_i$ are computed in parallel and aggregated by atomic addition into the corresponding $V_1$ and $V_2$.
After processing all pixels, the motion $[\hat A|\hat{\bm b}]$ of each Gaussian are computed by Equation~\ref{eq:weighted-least-squares} to obtain affine transformation $[A|\bm{b}]$ of each Gaussian.
To avoid numerical instability, we discard the motion of Gaussians that cover fewer than three pixels, or have a near-singular matrix $V_1$.

\subsection{Update Gaussians from Multi-view Observations}\label{sec:multi-view-motion}

Once we have motion $[A|\bm{b}]$ of each Gaussian in multiple views, we can compute its updated covariance matrix $\Sigma_{2D}'$ and 2D mean $\bm\mu_{2D}'$ according to Equation~\ref{eq:2d-transformation}, and then the updated 3D covariance matrix $\Sigma_{3D}'$ and 3D mean $\bm\mu_{3D}'$ can be derived from the multi-view 2D means and covariance.

\subsubsection{Compute 3D Mean}
Determining the 3D mean $\bm{\mu}_{3D}'$ from the 2D means $\bm{\mu}_{2D}'$ across multiple views is a typical triangulation problem.
With known camera intrinsic and extrinsic, at least two viewpoints are required to compute the $\bm{\mu}_{3D}'$ of a Gaussian.
We solve this triangulation problem using Singular Value Decomposition (SVD).
To maintain numerical stability, results are discarded for Gaussians observed in fewer than three views or those with low accumulated alpha values.

\subsubsection{Compute 3D Covariance Matrix}
Given the covariance matrices $\Sigma_{2D}'$ from multiple views and the parameters $J$ and $W$ of these views, the relationship $\Sigma_{2D}' = JW\Sigma_{3D}'W^\top J^\top$ in Equation~\ref{eq:3dgs} yields a linear system in $\Sigma_{3D}'$.
Each $\Sigma_{2D}'$ contributes three constraints, while $\Sigma_{3D}'$ has six unknown parameters.
Hence, at least two distinct views are required to solve for $\Sigma_{3D}'$.

\subsubsection{Decompose Covariance into Rotation and Scale}

According to Equation~\ref{eq:3dgs}, the 3D covariance matrix $\Sigma_{3D}$ comprises a rotation matrix $R$ and a scaling matrix $S$.
To update these parameters, we perform eigen decomposition on the modified covariance matrix $\Sigma_{3D}'$, extracting the updated rotation matrix $R'$ and scaling matrix $S'$.
In eigen decomposition, the eigenvector matrix corresponds to the rotation matrix, while the eigenvalues represent the squared scaling matrix.

However, naively using eigen decomposition can lead to two problems:
1) Negative eigenvalues cannot be square-rooted, making them unsuitable for computing the scaling matrix.
2) Eigenvalues are sorted from largest to smallest, which can disrupt the consistent order of rotation vectors and scaling factors between frames. This inconsistency can cause adjacent regions to appear to have similar relative rotations but significantly different rotation matrices, leading to instability in subsequent motion propagation.

To resolve these issues, we discard those $\Sigma_{3D}'$ with negative eigenvalues, and reorder the eigenvalues and corresponding eigenvectors to ensure their magnitudes match the order from the first frame.

\subsection{Motion Regularization}\label{sec:motion-regularization}
Pixel tracking and partial observations can introduce outlier Gaussians or wrong motions, leading to noticeable deviations.
We introduce a motion regularization to address these issues.
Specifically, we apply median filtering based on K-nearest neighbors to smooth the motion and then propagate the motion to the neighbor Gaussians that are not determined to be static.

\subsubsection{Median Filtering}\label{sec:median-filtering}

We observe that the majority of Gaussians are stable but a few are outliers.
This situation is similar to salt-and-pepper noise in image processing.
Therefore, we heuristically apply a median filtering approach.
Take $\Delta\mu_{3D}=\mu_{3D}'-\mu_{3D}$, $\Delta R=R'-R$, and $\Delta S=S'-S$ as the difference of a Gaussian between its first frame.
For each Gaussian, we find its K nearest neighbors in 3D space, and then compute the median of their $\Delta\mu_{3D}$, $\Delta R$, and $\Delta S$ and add these median values to the Gaussian parameters for the update.

\subsubsection{Propagation}\label{sec:propagation}

The motion of certain Gaussians may not be reliably determined, for example, due to low visibility (e.g., low opacity or visible in too few views) or negative eigenvalues arising from eigen decomposition.
For these Gaussians, we apply the motion propagation that propagates the motion from their neighbor to them.
Before propagation, we first figure out those static Gaussians by checking whether the pixels they cover are moving or not.
Specifically, we treat the pixels with the movement $|x_i'-x_i|$ given by the point tracking model as less than 1 pixel as static pixels, count them, and accumulate their alpha for each Gaussian in the rendering process.
The rule for detecting the static Gaussians can be varied.
In this paper, we take those Gaussians hit by more than 9 pixels and have more than 90\% of its pixels fixed in at least 2 views as static Gaussians, discard their computed motion, and exclude them from the motion propagation.
We then compute the average of their $\Delta\mu_{3D}$, $\Delta S$ and an average of rotations $\Delta R$ in Euler angle form and add these median values to the Gaussian parameters $\mu_{3D}$, $S$ and $R$.

\subsection{Refinement}\label{sec:refinement}

Even with point tracking and multi-view constraints, errors may persist in the recovered Gaussians.
We hence run a final refinement step by training the Gaussian parameters with the input video frames.
As the Gaussians have been moved to an approximately right position, the training process would be very fast and stable.

\subsection{Parallel Pipeline}\label{sec:parallel-pipeline}

\begin{figure}[htbp]
	\includegraphics{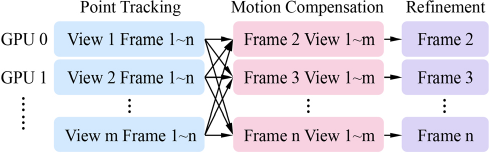}
    \caption{TrackerSplat parallel pipeline.}
    \label{fig:parallel}
\end{figure}

As illustrated in Figure~\ref{fig:parallel}, three key stages of TrackerSplat can be executed in parallel.
Since our point tracking operates per video clip, the point tracking can be parallelized by processing video from each view independently.
Additionally, the Motion Compensation and Refinement are both per-frame operations, they can be parallelized by processing each frame independently, only depending on the results of Point Tracking and initialization of the first frame.
The parallel pipeline can be implemented on multi-GPU systems and significantly improve the throughput of TrackerSplat.

\section{Experiments}

\begin{table*}[ht]
\caption{
    Quantitative comparison of average visual quality (\textit{PSNR} $\uparrow$ / \textit{SSIM} $\uparrow$ / \textit{LPIPs} $\downarrow$) in short-clip experiments under varying GPU parallelism settings (1, 2, 4, and 8 GPUs).
    Our method achieves better visual quality in most cases and demonstrates greater robustness than baselines as parallelism increases.
    Results for all other scenes are included in the supplementary material.
}
\label{tab:quality}
\setlength{\tabcolsep}{4pt}
\centering
\small
\begin{tabular}{l c | c c c c c c c}
\toprule
Method & GPUs & ``basketball'' & ``boxes'' & ``juggle'' & ``stepin'' & ``vrheadset'' & ``taekwondo'' & ``coffee martini'' \\
\midrule

P. ST-4DGS & 1 & 30.6 / .941 / .074 & 30.4 / .946 / .064 & 30.9 / .950 / .062 & 30.6 / .885 / .104 & 32.4 / .950 / .067 & 36.0 / .976 / .022 & 27.6 / \textbf{.918} / \textbf{.114} \\
P. 4DGS & 1 & 30.5 / .940 / \underline{.072} & 30.6 / .948 / .064 & 30.8 / .951 / .063 & 32.6 / .950 / .068 & 32.5 / .951 / .066 & 36.3 / .977 / .022 & 27.5 / .917 / .115 \\
P. Dy.3DGS & 1 & 30.9 / .942 / .075 & 31.5 / \underline{.951} / \underline{.061} & 31.7 / \underline{.953} / \underline{.061} & 32.4 / .948 / .070 & 32.4 / .951 / .067 & 36.6 / \underline{.978} / .021 & 27.6 / .917 / .115 \\
P. HiCoM & 1 & \underline{31.4} / \underline{.943} / .074 & \underline{31.6} / .949 / .065 & \underline{31.9} / .951 / .063 & \underline{32.9} / \underline{.951} / \underline{.067} & \underline{32.9} / \underline{.953} / \underline{.064} & \textbf{36.9} / \textbf{.979} / \textbf{.020} & \textbf{28.0} / \underline{.917} / .116 \\
Ours & 1 & \textbf{32.6} / \textbf{.950} / \textbf{.066} & \textbf{32.4} / \textbf{.953} / \textbf{.060} & \textbf{32.6} / \textbf{.956} / \textbf{.058} & \textbf{32.9} / \textbf{.952} / \textbf{.065} & \textbf{33.0} / \textbf{.954} / \textbf{.063} & \underline{36.8} / .978 / \underline{.020} & \underline{27.9} / .917 / \underline{.115} \\
\midrule
P. ST-4DGS & 2 & 30.2 / .936 / .081 & 30.4 / .944 / .066 & 31.0 / \underline{.951} / \underline{.063} & 32.1 / .947 / .071 & 32.2 / .950 / .067 & 35.0 / .971 / .027 & 27.6 / \textbf{.920} / \textbf{.110} \\
P. 4DGS & 2 & 30.7 / .940 / \underline{.073} & 30.2 / .941 / .068 & 30.6 / .947 / .066 & 32.6 / .950 / \underline{.068} & 32.4 / .950 / .066 & 35.5 / .974 / .024 & 27.4 / .918 / .113 \\
P. Dy.3DGS & 2 & 30.2 / .936 / .083 & 31.3 / \underline{.949} / \underline{.063} & 31.4 / .951 / .064 & 32.2 / .947 / .072 & 32.0 / .950 / .067 & 36.4 / .977 / .023 & 27.6 / .918 / .113 \\
P. HiCoM & 2 & \underline{31.0} / \underline{.940} / .080 & \underline{31.6} / .948 / .066 & \underline{31.9} / .951 / .065 & \underline{32.8} / \underline{.951} / .068 & \underline{32.8} / \underline{.953} / \underline{.065} & \underline{36.7} / \textbf{.978} / \underline{.021} & \textbf{28.0} / \underline{.919} / .113 \\
Ours & 2 & \textbf{32.6} / \textbf{.950} / \textbf{.067} & \textbf{32.2} / \textbf{.953} / \textbf{.061} & \textbf{32.7} / \textbf{.956} / \textbf{.059} & \textbf{32.9} / \textbf{.953} / \textbf{.065} & \textbf{33.0} / \textbf{.954} / \textbf{.062} & \textbf{36.8} / \underline{.977} / \textbf{.021} & \underline{27.9} / .919 / \underline{.112} \\
\midrule
P. ST-4DGS & 4 & 29.7 / .932 / .089 & 30.2 / .943 / .069 & 31.0 / .949 / \underline{.066} & 31.9 / .946 / .074 & 31.9 / .949 / .068 & 28.4 / .778 / .172 & 27.5 / .919 / \textbf{.111} \\
P. 4DGS & 4 & 30.0 / .932 / \underline{.082} & 30.1 / .944 / .068 & 30.9 / \underline{.950} / .066 & 32.6 / \underline{.950} / \underline{.068} & 32.0 / .949 / .068 & 34.8 / .972 / .028 & 27.4 / .917 / .114 \\
P. Dy.3DGS & 4 & 29.1 / .927 / .096 & 30.8 / .946 / \underline{.068} & 30.8 / .946 / .071 & 31.7 / .945 / .075 & 31.5 / .948 / .070 & 35.8 / .975 / .026 & 27.6 / .918 / .114 \\
P. HiCoM & 4 & \underline{30.4} / \underline{.936} / .088 & \underline{31.4} / \underline{.946} / .069 & \underline{31.6} / .949 / .068 & \underline{32.6} / .950 / .070 & \underline{32.6} / \underline{.952} / \underline{.066} & \underline{36.4} / \underline{.977} / \underline{.022} & \textbf{28.0} / \underline{.919} / .114 \\
Ours & 4 & \textbf{32.4} / \textbf{.948} / \textbf{.070} & \textbf{32.4} / \textbf{.953} / \textbf{.062} & \textbf{32.6} / \textbf{.955} / \textbf{.060} & \textbf{32.8} / \textbf{.952} / \textbf{.066} & \textbf{32.9} / \textbf{.954} / \textbf{.063} & \textbf{36.7} / \textbf{.977} / \textbf{.021} & \underline{27.9} / \textbf{.919} / \underline{.112} \\
\midrule
P. ST-4DGS & 8 & 29.1 / .924 / .099 & 30.0 / .941 / \underline{.073} & 30.6 / \underline{.947} / \underline{.069} & 30.1 / .883 / .108 & 31.6 / .947 / .071 & 33.7 / .968 / .033 & 27.4 / .916 / \underline{.113} \\
P. 4DGS & 8 & 29.6 / \underline{.928} / \underline{.091} & 29.7 / .937 / .076 & 30.2 / .940 / .071 & \underline{32.4} / \underline{.949} / \underline{.069} & 31.6 / .947 / .070 & 33.8 / .966 / .035 & 27.3 / .916 / .115 \\
P. Dy.3DGS & 8 & 27.7 / .915 / .117 & 30.1 / .942 / .074 & 29.7 / .938 / .082 & 31.2 / .943 / .079 & 30.8 / .945 / .073 & 35.0 / .971 / .032 & 27.5 / .917 / .115 \\
P. HiCoM & 8 & \underline{29.6} / .928 / .102 & \underline{31.0} / \underline{.943} / .074 & \underline{31.3} / .945 / .075 & 32.2 / .948 / .075 & \underline{32.1} / \underline{.950} / \underline{.069} & \underline{36.0} / \underline{.975} / \underline{.026} & \textbf{28.0} / \underline{.918} / .115 \\
Ours & 8 & \textbf{31.9} / \textbf{.944} / \textbf{.076} & \textbf{31.9} / \textbf{.951} / \textbf{.064} & \textbf{32.5} / \textbf{.954} / \textbf{.062} & \textbf{32.7} / \textbf{.952} / \textbf{.067} & \textbf{32.7} / \textbf{.953} / \textbf{.064} & \textbf{36.5} / \textbf{.976} / \textbf{.022} & \underline{27.9} / \textbf{.919} / \textbf{.112} \\

\bottomrule
\end{tabular}
\end{table*}

\subsection{Datasets}

We evaluate our method on four widely-used dynamic scene datasets:

\textbf{Meeting Room} dataset~\cite{liStreamingRadianceFields2022} includes 4 scenes, each of 300 frames captured from 13 viewpoints at a resolution of 1280$\times$720.

\textbf{Neural 3D Video Synthesis} (\textbf{N3DV}) dataset~\cite{liNeural3DVideo2022} contains 6 scenes, each of 300 frames captured from 18 views at 2704$\times$2028 resolution.

\textbf{Dynamic3DGS} dataset~\cite{luiten2023dynamic} comprise 6 scenes (150 frames each) captured from 27 views at 640$\times$360 resolution.

\textbf{st-nerf} dataset~\cite{zhangEditableFreeviewpointVideo2021} consists of 3 scenes, each contains 75--100 frames from 15 views at 1920$\times$1080 resolution.

\textbf{RH20T} dataset~\cite{fangRH20TComprehensiveRobotic2024} includes 110k robotic manipulation sequences from 8-10 views at 640x360 resolution.
For evaluation, we selected 14 videos and extracted the longest continuous multi-view segments synchronized within 100ms across views.

We use these datasets in two experiment setups:
\begin{itemize}
\item \textbf{Short-clip experiments}: 
Videos are segmented into clips of 2, 3, 5, or 9 frames (without overlapping).
The first frame of each clip is reconstructed in the initialization stage, and subsequent frames are processed in parallel using 1, 2, 4, or 8 GPUs, respectively.
This setup evaluates quality degradation caused solely by increased parallelism without cumulative errors across clips.

\item \textbf{Long-video experiments}: 
Clips are sequentially connected, with the last frame of each clip serving as the first frame of the next.
Only the first frame of the first clip is reconstructed in the initialization stage.
This setup evaluates robustness and temporal consistency over longer sequences.
\end{itemize}

\subsection{Implementation Details}

We carefully evaluated various hyperparameter settings and selected those achieving the best balance between quality and efficiency.

\paragraph{Initialization}
Following InstantSplat~\cite{fanInstantSplatUnboundedSparseview2024}, we initialize Gaussians from COLMAP point clouds and optimize Gaussian parameters for 10,000 iterations, providing a stable starting point for subsequent frames.

\paragraph{Point Tracking}
We evaluated DOT~\cite{lemoingDenseOpticalTracking2024}, CoTracker~\cite{karaevCoTrackerItBetter2025,karaevCoTracker3SimplerBetter2024}, and TAPIR~\cite{doerschTAPIRTrackingAny2023}.
DOT provides similar accuracy at much higher speed and was therefore selected.
To improve efficiency, input images are resized (Meeting Room \& Dynamic3DGS: 640$\times$360, N3DV: 800$\times$600, st-nerf: 960$\times$540).

\paragraph{Regularization}
Gaussians are marked unsolvable in PWI-LS if their $V_1$ determinant is below $10^{-12}$, their accumulated alpha is below $10^{-3}$, or they cover fewer than 2 pixels.
Gaussians are marked unsolvable during multi-view updates if visible in fewer than 2 views, accumulated alpha is below $10^{-3}$, or covering fewer than 3 pixels across all views.
In the regularization stage, parameters of unsolvable Gaussians are updated based on their 8 nearest neighbors.

\paragraph{Refinement}
In the refinement stage, Gaussian parameters are optimized following the original 3DGS method~\cite{kerbl3DGaussianSplatting2023} for 1,000 iterations without densification.

\paragraph{Runtime}
Our implementation is built on PyTorch, Taichi~\cite{hu2019taichi}, and CUDA.
All experiments were conducted on a server with 8 NVIDIA A100-SXM4-40GB GPUs.

\subsection{Baseline and Ablation}

We compare TrackerSplat with representative dynamic scene reconstruction methods that support Gaussian trajectory tracking from multi-view videos with fixed camera poses.
The original implementations of these methods do not directly support multi-GPU parallel processing.
Thus, to ensure a fair comparison, we carefully copy and adapt their codebases to our parallel framework, strictly preserving their original designs and hyperparameters to minimize any deviation from their original implementations:

\paragraph{Parallel HiCoM}
Hierarchical Coherent Motion (HiCoM)~\cite{gaoHiCoMHierarchicalCoherent2024} associates Gaussian motions with distinct regions.
In their open-source code, HiCoM is implemented as a hierarchical grid with multiple density levels.
We copy and adapt this HiCoM grid implementation to our parallel framework.

\paragraph{Parallel Dynamic 3DGS}
Dynamic 3DGS~\cite{luiten2023dynamic} sequentially train Gaussian parameters frame-by-frame with physics-based regularization.
We copy their regularization term and adapt their sequential training to parallel processing by initializing the training from the first frame of each clip rather than previous frame.

\paragraph{Parallel 4DGS}
4DGS~\cite{wu4DGaussianSplatting2024} introduces a deformation field to represent Gaussian motion.
We copy and adapt the deformation field implementation for parallel processing by training a separate deformation field per frame, warping Gaussians from the initial frame to subsequent frames within each clip.

\paragraph{Parallel ST-4DGS}
ST-4DGS~\cite{liST4DGSSpatialTemporallyConsistent2024} extends 4DGS by incorporating explicit temporal regularization.
We copy and adapt their regularization term into our 4DGS implementation.

\paragraph{TrackerSplat without regularization (Ablation)}
As an ablation study, we remove the regularization stage and directly trained the Gaussian parameters based on the motion compensation results.

We observed that the regularization in ST-4DGS, 4DGS, and Dynamic 3DGS, originally tuned for small motions, tends to over-constrain Gaussians in our parallel setting, producing severe artifacts (e.g., sticking and fading) that obscure the true performance of these baselines.
To ensure a fair comparison, we reduce their regularization weights, which significantly alleviates these artifacts.
HiCoM is less affected because its regularization is weaker and more localized.
Moreover, since our refinement stage is essentially training without regularization, including HiCoM in our evaluation highlights that the robustness of TrackerSplat is not solely due to the absence of regularization during refinement.
All baselines share the same initial frame for both short-clip and long-video experiments.

\subsection{Comparison of Visual Quality}

We evaluate rendering quality using three widely accepted metrics: structural similarity (SSIM), peak signal-to-noise ratio (PSNR), and perceptual similarity (LPIPS)~\cite{zhang2018perceptual}.

\paragraph{Visual quality in short clips}
Table~\ref{tab:quality} presents quantitative results from short-clip experiments under different GPU parallelism settings.
In most cases, our method outperforms baselines in visual quality on single GPUs and maintains higher quality as parallelism increases.
This demonstrates robustness of TrackerSplat across different parallelism settings.
In scenes with slow motion (e.g., "coffee martini"), baselines experience minor quality drops (around 0.1 PSNR), and achieve performance comparable to ours.
Figure~\ref{fig:rendering} provides visual comparisons, illustrating that baselines often suffer from fading or drifting artifacts in fast-moving regions, while our method better preserves visual fidelity, even in highly dynamic scenes.
This demonstrates that explicit motion compensation prior to training effectively mitigates the impact of large displacements, maintaining high quality across different levels of parallelism.

\paragraph{Visual quality in long videos}
Figure~\ref{fig:framequality-full} shows quantitative results over long sequences.
Our method achieves higher and more stable visual quality compared to baselines in most cases.
Sample rendered videos are provided in the supplementary material.

\subsection{Ablation Study Results}
\begin{figure*}[htbp]
	\centering
	\includegraphics{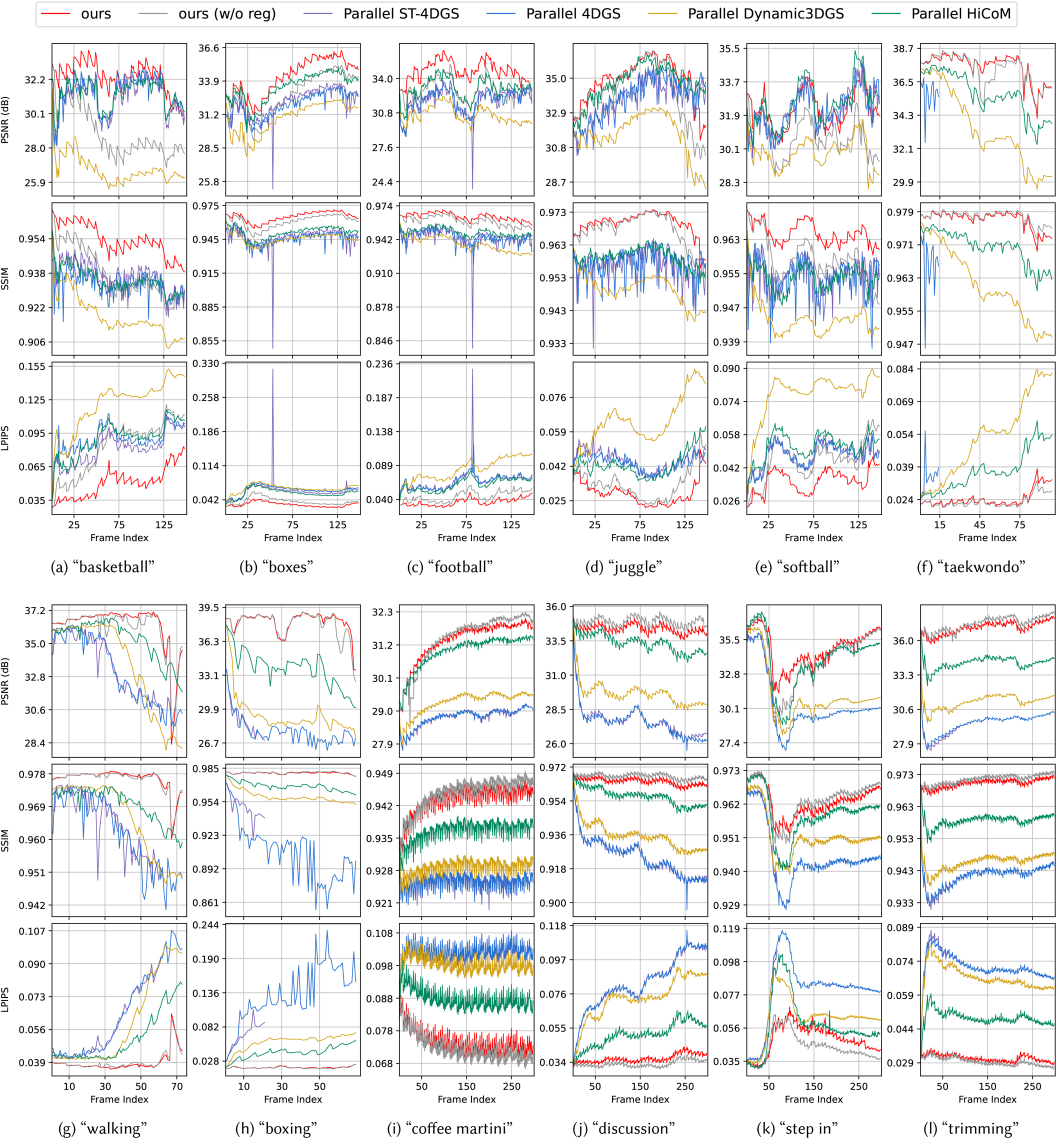}
	\caption{
		Average visual quality (\textit{PSNR} $\uparrow$ / \textit{SSIM} $\uparrow$ / \textit{LPIPs} $\downarrow$) over long-video sequences using our parallel pipeline with 8 GPUs (long-video experiments).
		Our method achieves higher and more stable visual quality than baselines in most cases, demonstrating its robustness.
		Lines ending prematurely for 4DGS and ST-4DGS indicate training failures due to GPU memory overflow (exceeding the 40GB limit of the A100 GPU) or numerical instabilities (NaN gradients).
		Corresponding rendered videos are provided in the supplementary material.
	}\label{fig:framequality-full}
\end{figure*}

Figure~\ref{fig:framequality-full} compares TrackerSplat with and without motion regularization.
Motion regularization generally improves robustness but can slightly reduce quality in certain cases (e.g., "coffee martini").
Profiling reveals that regularization effectively corrects outliers but may also shift already well-positioned Gaussians, slightly degrading quality.
For short, robustness of our method against point tracking error stems from several design choices:
1) Point tracker provides accurate pixel trajectories in most cases.
2) PWI-LS (Sec.~\ref{sec:parallel-weighted-incremental-least-squares}) mitigates individual pixel tracking errors by computing transparency-weighted averages.
3) Multi-view triangulation (Sec.~\ref{sec:multi-view-motion}) corrects trajectory errors by averaging results from multiple viewpoints.
4) Median filtering (Sec.\ref{sec:median-filtering}) and propagation (Sec.\ref{sec:propagation}) handle severe trajectory errors by replacing incorrect trajectories with neighboring Gaussians.
5) The refinement stage (Sec.~\ref{sec:refinement}) further optimizes Gaussian positions, enhancing final accuracy.

\subsection{Comparison of Parallel Performance}
Table~\ref{tab:speed} compares parallel performance across different GPU settings.
TrackerSplat achieves higher throughput in most settings.
Our parallel framework can also improve the throughput of existing methods in multi-GPU environments.
Baselines are generally slower due to the computational demands of their regularization terms or deformation fields.
In contrast, although includes additional tracking stages, our method benefits from accurately aligned Gaussians before training, eliminating the need for complex regularization or deformation fields, and thus improving reconstruction throughput.
Dynamic3DGS dataset contains more views (27) but at a lower resolution (only 640$\times$360).
In this dataset, the increased number of views introduces additional overhead in the point tracking stage.
Furthermore, since all methods perform the same number of training/refinement iterations (1,000), the overhead from tracking outweighs the performance gains obtained from simplified training.
As a result, our method shows comparatively lower throughput than baselines on this dataset.

\begin{table}[ht]
\caption{
    Average throughput comparison (seconds per frame) between TrackerSplat and baselines under different GPU parallelism settings (1, 2, 4, and 8 GPUs).
    Our method achieves the highest throughput in most cases.
    We also separately report the runtime of the tracking stages (point tracking and motion compensation) and the refinement stage.
}
\setlength{\tabcolsep}{4pt}
\centering
\small
\begin{tabular}{l c | c c c c }
\toprule
Method & GPUs & Dy.3DGS & Meet.Room & st-nerf & N3DV \\
\midrule
Parallel ST-4DGS & 1 & 217.8 & 344.7 & 361.9 & 464.2 \\
Parallel 4DGS & 1 & 52.3 & 84.9 & 109.9 & 168.0 \\
Parallel Dyn.3DGS & 1 & 31.7 & 42.1 & 72.1 & 123.3 \\
Parallel HiCoM & 1 & \underline{13.1} & \underline{26.9} & \underline{54.1} & \underline{119.0} \\
Ours (total) & 1 & \textbf{13.0} & \textbf{20.1} & \textbf{51.3} & \textbf{109.3} \\
Ours (track+refine) & 1 & 5.3+7.6 & 3.3+16.9 & 6.3+45.1 & 10.0+99.3 \\
\midrule
Parallel ST-4DGS & 2 & 108.9 & 172.3 & 181.0 & 232.1 \\
Parallel 4DGS & 2 & 26.1 & 42.4 & 54.9 & 84.0 \\
Parallel Dyn.3DGS & 2 & 15.8 & 21.1 & 36.1 & 61.6 \\
Parallel HiCoM & 2 & \textbf{6.6} & \underline{13.4} & \underline{27.0} & \underline{59.5} \\
Ours (total) & 2 & \underline{7.7} & \textbf{10.7} & \textbf{26.6} & \textbf{55.5} \\
Ours (track+refine) & 2 & 3.9+3.8 & 2.3+8.4 & 4.1+22.5 & 5.9+49.6 \\
\midrule
Parallel ST-4DGS & 4 & 54.4 & 86.2 & 90.5 & 116.0 \\
Parallel 4DGS & 4 & 13.1 & 21.2 & 27.5 & 42.0 \\
Parallel Dyn.3DGS & 4 & 7.9 & 10.5 & 18.0 & 30.8 \\
Parallel HiCoM & 4 & \textbf{3.3} & \underline{6.7} & \underline{14.8} & \underline{29.7} \\
Ours (total) & 4 & \underline{5.2} & \textbf{6.0} & \textbf{13.5} & \textbf{28.1} \\
Ours (track+refine) & 4 & 3.3+1.9 & 1.7+4.2 & 2.2+11.3 & 3.3+24.8 \\
\midrule
Parallel ST-4DGS & 8 & 27.2 & 43.1 & 45.2 & 58.0 \\
Parallel 4DGS & 8 & 6.5 & 10.6 & 13.7 & 21.0 \\
Parallel Dyn.3DGS & 8 & 4.0 & 5.3 & 9.0 & 15.4 \\
Parallel HiCoM & 8 & \textbf{1.6} & \underline{3.4} & \underline{7.4} & \underline{14.9} \\
Ours (total) & 8 & \underline{3.4} & \textbf{3.4} & \textbf{7.1} & \textbf{14.7} \\
Ours (track+refine) & 8 & 2.4+1.0 & 1.3+2.1 & 1.5+5.6 & 2.3+12.4 \\
\bottomrule
\end{tabular}
\label{tab:speed}
\end{table}

\section{Limitations and Future Work}

\begin{figure*}[htbp]
	\centering
	\includegraphics{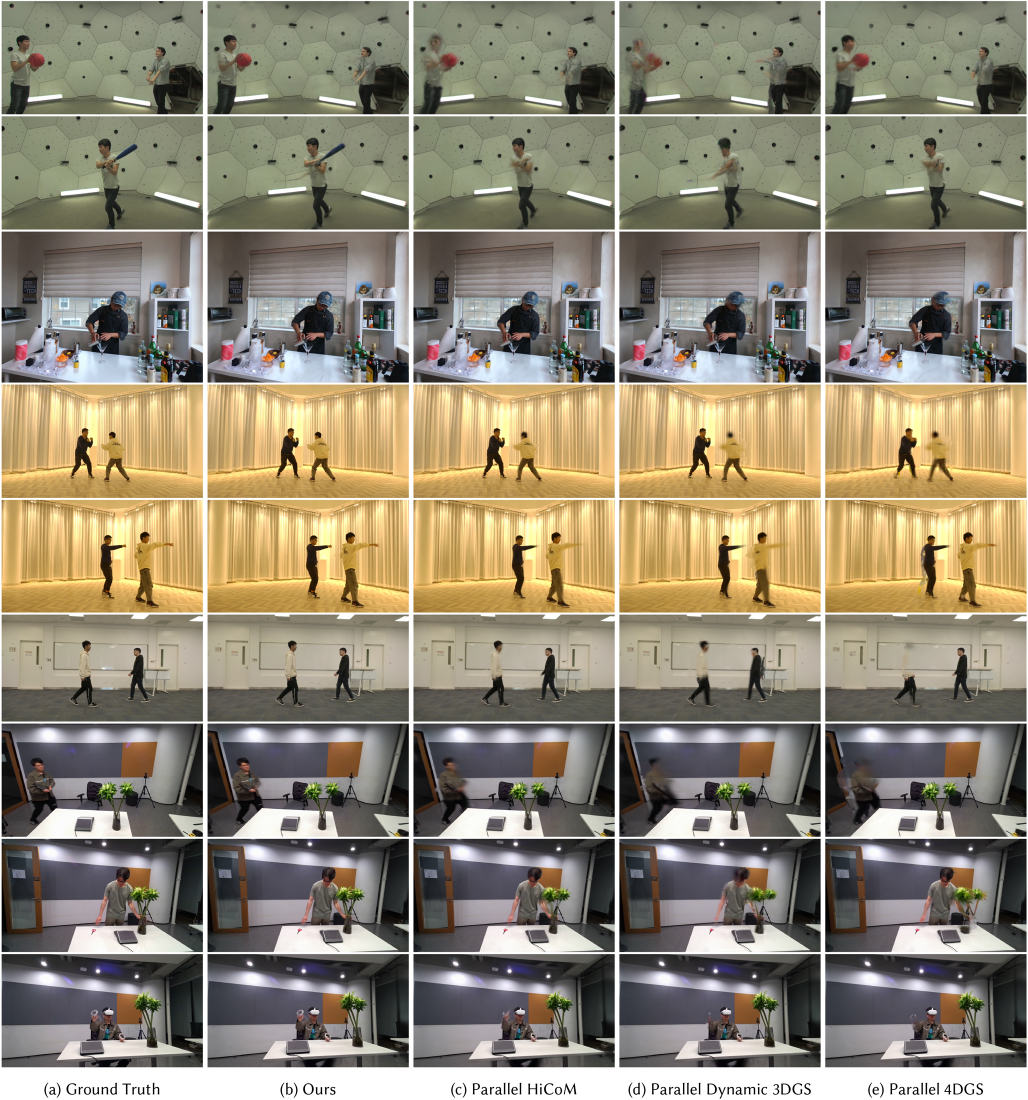}
	\caption{
		Qualitative comparison of rendered results from the final frame of representative 9-frame clips processed in parallel using 8 GPUs (short-clip experiments).
		Our method generates fewer artifacts and better preserves visual details compared to baselines, particularly in highly dynamic regions.
	}\label{fig:rendering}
\end{figure*}

Although TrackerSplat address the  limitations in position gradient for large inter-frame displacements, it still has certain limitations:

\paragraph{Small or thin objects}
Point trackers often miss subtle motions in small or thin objects (e.g., human hands or fingers shown in Figure~\ref{fig:rendering}), leading to blurred or missing details.

\paragraph{Jittering artifacts}
We observe temporal jitter (see the supplementary video) from two sources:
1) tracking errors, especially in low-texture regions where correspondence is ambiguous; although Motion Regularization (Sec.\ref{sec:motion-regularization}) and Refinement (Sec.\ref{sec:refinement}) mitigate these issues, they do not eliminate them;
and 2) spatiotemporal inconsistencies of the 3DGS representation under noisy video data \cite{yunCompensatingSpatiotemporallyInconsistent2025}.

\paragraph{Accumulated errors}
While the refinement step can correct some errors from earlier frames, it is not sufficiently robust.
As a result, inaccuracies may accumulate over time, and errors in the first frame can propagate to subsequent frames, especially in challenging cases with limited training views or complex scenes.

\paragraph{Occlusions}
Typical point tracking methods establish trajectories by matching pixels in subsequent frames with those in the first input frame.
Therefore, if an object is fully occluded in the first frame and becomes visible later in a clip, point tracker may fail to estimate its trajectory correctly, causing incomplete or missing reconstructions.

\paragraph{Potential Solutions}
Recent studies~\cite{duan4DRotorGaussianSplatting2024,sun3DGStreamFlyTraining2024} propose techniques that dynamically remove faded Gaussians and add new ones in regions with high gradients, potentially addressing these limitations.
As these methods do not inherently maintain consistent Gaussian trajectories across frames,
integrating such techniques with TrackerSplat would require additional mechanisms to match newly added Gaussians to existing trajectories.
Moreover, error decomposition for 3DGS~\cite{yunCompensatingSpatiotemporallyInconsistent2025} directly targets spatiotemporal inconsistency and could be incorporated into our refinement stage.
Exploring this integration represents a promising direction for future research.

\section{Conclusion}

We presented TrackerSplat, a robust and scalable approach for dynamic scene reconstruction using 3DGS.
TrackerSplat integrates off-the-shelf point tracker to extract pixel trajectories, and triangulates them across views to update Gaussians before refinement.
This design enables TrackerSplat to effectively manage large inter-frame motions, substantially improving reconstruction throughput in multi-GPU environments while maintaining high visual quality.
Evaluation results on real-world dynamic scenes demonstrate the effectiveness and scalability of TrackerSplat, paving the way toward real-time dynamic scene reconstruction.

\begin{acks}
This research is supported by an NSERC Discovery Grant and a MITACS Accelerate Cluster Grant.
\end{acks}

\bibliographystyle{ACM-Reference-Format}

\begin{thebibliography}{28}


\ifx \showCODEN    \undefined \def \showCODEN     #1{\unskip}     \fi
\ifx \showISBNx    \undefined \def \showISBNx     #1{\unskip}     \fi
\ifx \showISBNxiii \undefined \def \showISBNxiii  #1{\unskip}     \fi
\ifx \showISSN     \undefined \def \showISSN      #1{\unskip}     \fi
\ifx \showLCCN     \undefined \def \showLCCN      #1{\unskip}     \fi
\ifx \shownote     \undefined \def \shownote      #1{#1}          \fi
\ifx \showarticletitle \undefined \def \showarticletitle #1{#1}   \fi
\ifx \showURL      \undefined \def \showURL       {\relax}        \fi
\providecommand\bibfield[2]{#2}
\providecommand\bibinfo[2]{#2}
\providecommand\natexlab[1]{#1}
\providecommand\showeprint[2][]{arXiv:#2}

\bibitem[{Abou-Chakra} et~al\mbox{.}(2024)]%
        {abou-chakraPhysicallyEmbodiedGaussian2024}
\bibfield{author}{\bibinfo{person}{Jad {Abou-Chakra}}, \bibinfo{person}{Krishan
  Rana}, \bibinfo{person}{Feras Dayoub}, {and} \bibinfo{person}{Niko
  Suenderhauf}.} \bibinfo{year}{2024}\natexlab{}.
\newblock \showarticletitle{Physically {{Embodied Gaussian Splatting}}: {{A
  Visually Learnt}} and {{Physically Grounded 3D Representation}} for
  {{Robotics}}}. In \bibinfo{booktitle}{\emph{8th {{Annual Conference}} on
  {{Robot Learning}}}}.
\newblock


\bibitem[Doersch et~al\mbox{.}(2023)]%
        {doerschTAPIRTrackingAny2023}
\bibfield{author}{\bibinfo{person}{Carl Doersch}, \bibinfo{person}{Yi Yang},
  \bibinfo{person}{Mel Vecerik}, \bibinfo{person}{Dilara Gokay},
  \bibinfo{person}{Ankush Gupta}, \bibinfo{person}{Yusuf Aytar},
  \bibinfo{person}{Joao Carreira}, {and} \bibinfo{person}{Andrew Zisserman}.}
  \bibinfo{year}{2023}\natexlab{}.
\newblock \showarticletitle{{{TAPIR}}: {{Tracking Any Point}} with {{Per-Frame
  Initialization}} and {{Temporal Refinement}}}. In
  \bibinfo{booktitle}{\emph{Proceedings of the {{IEEE}}/{{CVF International
  Conference}} on {{Computer Vision}}}}. \bibinfo{pages}{10061--10072}.
\newblock


\bibitem[Duan et~al\mbox{.}(2024)]%
        {duan4DRotorGaussianSplatting2024}
\bibfield{author}{\bibinfo{person}{Yuanxing Duan}, \bibinfo{person}{Fangyin
  Wei}, \bibinfo{person}{Qiyu Dai}, \bibinfo{person}{Yuhang He},
  \bibinfo{person}{Wenzheng Chen}, {and} \bibinfo{person}{Baoquan Chen}.}
  \bibinfo{year}{2024}\natexlab{}.
\newblock \showarticletitle{{{4D-Rotor Gaussian Splatting}}: {{Towards
  Efficient Novel View Synthesis}} for {{Dynamic Scenes}}}. In
  \bibinfo{booktitle}{\emph{{{ACM SIGGRAPH}} 2024 {{Conference Papers}}}}
  \emph{(\bibinfo{series}{{{SIGGRAPH}} '24})}. \bibinfo{pages}{1--11}.
\newblock
\showISBNx{979-8-4007-0525-0}


\bibitem[Fan et~al\mbox{.}(2024)]%
        {fanInstantSplatUnboundedSparseview2024}
\bibfield{author}{\bibinfo{person}{Zhiwen Fan}, \bibinfo{person}{Wenyan Cong},
  \bibinfo{person}{Kairun Wen}, \bibinfo{person}{Kevin Wang},
  \bibinfo{person}{Jian Zhang}, \bibinfo{person}{Xinghao Ding},
  \bibinfo{person}{Danfei Xu}, \bibinfo{person}{Boris Ivanovic},
  \bibinfo{person}{Marco Pavone}, \bibinfo{person}{Georgios Pavlakos},
  \bibinfo{person}{Zhangyang Wang}, {and} \bibinfo{person}{Yue Wang}.}
  \bibinfo{year}{2024}\natexlab{}.
\newblock \bibinfo{title}{{{InstantSplat}}: {{Unbounded Sparse-view Pose-free
  Gaussian Splatting}} in 40 {{Seconds}}}.
\newblock
\href{https://doi.org/10.48550/ARXIV.2403.20309}{doi:\nolinkurl{10.48550/ARXIV.2403.20309}}


\bibitem[Fang et~al\mbox{.}(2024)]%
        {fangRH20TComprehensiveRobotic2024}
\bibfield{author}{\bibinfo{person}{Hao-Shu Fang}, \bibinfo{person}{Hongjie
  Fang}, \bibinfo{person}{Zhenyu Tang}, \bibinfo{person}{Jirong Liu},
  \bibinfo{person}{Chenxi Wang}, \bibinfo{person}{Junbo Wang},
  \bibinfo{person}{Haoyi Zhu}, {and} \bibinfo{person}{Cewu Lu}.}
  \bibinfo{year}{2024}\natexlab{}.
\newblock \showarticletitle{{{RH20T}}: {{A Comprehensive Robotic Dataset}} for
  {{Learning Diverse Skills}} in {{One-Shot}}}. In
  \bibinfo{booktitle}{\emph{2024 {{IEEE International Conference}} on
  {{Robotics}} and {{Automation}} ({{ICRA}})}}. \bibinfo{pages}{653--660}.
\newblock


\bibitem[Gao et~al\mbox{.}(2024)]%
        {gaoHiCoMHierarchicalCoherent2024}
\bibfield{author}{\bibinfo{person}{Qiankun Gao}, \bibinfo{person}{Jiarui Meng},
  \bibinfo{person}{Chengxiang Wen}, \bibinfo{person}{Jie Chen}, {and}
  \bibinfo{person}{Jian Zhang}.} \bibinfo{year}{2024}\natexlab{}.
\newblock \showarticletitle{{{HiCoM}}: {{Hierarchical Coherent Motion}} for
  {{Dynamic Streamable Scenes}} with {{3D Gaussian Splatting}}}. In
  \bibinfo{booktitle}{\emph{The {{Thirty-eighth Annual Conference}} on {{Neural
  Information Processing Systems}}}}.
\newblock


\bibitem[Hu et~al\mbox{.}(2019)]%
        {hu2019taichi}
\bibfield{author}{\bibinfo{person}{Yuanming Hu}, \bibinfo{person}{Tzu-Mao Li},
  \bibinfo{person}{Luke Anderson}, \bibinfo{person}{Jonathan Ragan-Kelley},
  {and} \bibinfo{person}{Fr{\'e}do Durand}.} \bibinfo{year}{2019}\natexlab{}.
\newblock \showarticletitle{Taichi: a language for high-performance computation
  on spatially sparse data structures}.
\newblock \bibinfo{journal}{\emph{ACM Transactions on Graphics (TOG)}}
  \bibinfo{volume}{38}, \bibinfo{number}{6} (\bibinfo{year}{2019}),
  \bibinfo{pages}{201}.
\newblock


\bibitem[Karaev et~al\mbox{.}(2024)]%
        {karaevCoTracker3SimplerBetter2024}
\bibfield{author}{\bibinfo{person}{Nikita Karaev}, \bibinfo{person}{Iurii
  Makarov}, \bibinfo{person}{Jianyuan Wang}, \bibinfo{person}{Natalia
  Neverova}, \bibinfo{person}{Andrea Vedaldi}, {and} \bibinfo{person}{Christian
  Rupprecht}.} \bibinfo{year}{2024}\natexlab{}.
\newblock \bibinfo{title}{{{CoTracker3}}: {{Simpler}} and {{Better Point
  Tracking}} by {{Pseudo-Labelling Real Videos}}}.
\newblock
\href{https://doi.org/10.48550/ARXIV.2410.11831}{doi:\nolinkurl{10.48550/ARXIV.2410.11831}}


\bibitem[Karaev et~al\mbox{.}(2025)]%
        {karaevCoTrackerItBetter2025}
\bibfield{author}{\bibinfo{person}{Nikita Karaev}, \bibinfo{person}{Ignacio
  Rocco}, \bibinfo{person}{Benjamin Graham}, \bibinfo{person}{Natalia
  Neverova}, \bibinfo{person}{Andrea Vedaldi}, {and} \bibinfo{person}{Christian
  Rupprecht}.} \bibinfo{year}{2025}\natexlab{}.
\newblock \showarticletitle{{{CoTracker}}: {{It Is Better}} to~{{Track
  Together}}}. In \bibinfo{booktitle}{\emph{Computer {{Vision}} -- {{ECCV}}
  2024}}, \bibfield{editor}{\bibinfo{person}{Ale{\v s} Leonardis},
  \bibinfo{person}{Elisa Ricci}, \bibinfo{person}{Stefan Roth},
  \bibinfo{person}{Olga Russakovsky}, \bibinfo{person}{Torsten Sattler}, {and}
  \bibinfo{person}{G{\"u}l Varol}} (Eds.). \bibinfo{pages}{18--35}.
\newblock
\showISBNx{978-3-031-73033-7}


\bibitem[Kerbl et~al\mbox{.}(2023)]%
        {kerbl3DGaussianSplatting2023}
\bibfield{author}{\bibinfo{person}{Bernhard Kerbl}, \bibinfo{person}{Georgios
  Kopanas}, \bibinfo{person}{Thomas Leimkuehler}, {and} \bibinfo{person}{George
  Drettakis}.} \bibinfo{year}{2023}\natexlab{}.
\newblock \showarticletitle{{{3D Gaussian Splatting}} for {{Real-Time Radiance
  Field Rendering}}}.
\newblock \bibinfo{journal}{\emph{ACM Transactions on Graphics}}
  \bibinfo{volume}{42}, \bibinfo{number}{4} (\bibinfo{year}{2023}),
  \bibinfo{pages}{139:1--139:14}.
\newblock
\showISSN{0730-0301}


\bibitem[Le~Moing et~al\mbox{.}(2024)]%
        {lemoingDenseOpticalTracking2024}
\bibfield{author}{\bibinfo{person}{Guillaume Le~Moing}, \bibinfo{person}{Jean
  Ponce}, {and} \bibinfo{person}{Cordelia Schmid}.}
  \bibinfo{year}{2024}\natexlab{}.
\newblock \showarticletitle{Dense {{Optical Tracking}}: {{Connecting}} the
  {{Dots}}}. In \bibinfo{booktitle}{\emph{Proceedings of the {{IEEE}}/{{CVF
  Conference}} on {{Computer Vision}} and {{Pattern Recognition}}}}.
  \bibinfo{pages}{19187--19197}.
\newblock


\bibitem[Lei et~al\mbox{.}(2025)]%
        {leiMoScaDynamicGaussian2025}
\bibfield{author}{\bibinfo{person}{Jiahui Lei}, \bibinfo{person}{Yijia Weng},
  \bibinfo{person}{Adam~W. Harley}, \bibinfo{person}{Leonidas Guibas}, {and}
  \bibinfo{person}{Kostas Daniilidis}.} \bibinfo{year}{2025}\natexlab{}.
\newblock \showarticletitle{{{MoSca}}: {{Dynamic Gaussian Fusion}} from
  {{Casual Videos}} via {{4D Motion Scaffolds}}}. In
  \bibinfo{booktitle}{\emph{Proceedings of the {{IEEE}}/{{CVF Conference}} on
  {{Computer Vision}} and {{Pattern Recognition}}}}.
\newblock


\bibitem[Li et~al\mbox{.}(2024b)]%
        {liST4DGSSpatialTemporallyConsistent2024}
\bibfield{author}{\bibinfo{person}{Deqi Li}, \bibinfo{person}{Shi-Sheng Huang},
  \bibinfo{person}{Zhiyuan Lu}, \bibinfo{person}{Xinran Duan}, {and}
  \bibinfo{person}{Hua Huang}.} \bibinfo{year}{2024}\natexlab{b}.
\newblock \showarticletitle{{{ST-4DGS}}: {{Spatial-Temporally Consistent 4D
  Gaussian Splatting}} for {{Efficient Dynamic Scene Rendering}}}. In
  \bibinfo{booktitle}{\emph{{{ACM SIGGRAPH}} 2024 {{Conference Papers}}}}
  \emph{(\bibinfo{series}{{{SIGGRAPH}} '24})}. \bibinfo{pages}{1--11}.
\newblock
\showISBNx{979-8-4007-0525-0}


\bibitem[Li et~al\mbox{.}(2022a)]%
        {liStreamingRadianceFields2022}
\bibfield{author}{\bibinfo{person}{Lingzhi Li}, \bibinfo{person}{Zhen Shen},
  \bibinfo{person}{Zhongshu Wang}, \bibinfo{person}{Li Shen}, {and}
  \bibinfo{person}{Ping Tan}.} \bibinfo{year}{2022}\natexlab{a}.
\newblock \showarticletitle{Streaming {{Radiance Fields}} for {{3D Video
  Synthesis}}}.
\newblock \bibinfo{journal}{\emph{Advances in Neural Information Processing
  Systems}}  \bibinfo{volume}{35} (\bibinfo{year}{2022}),
  \bibinfo{pages}{13485--13498}.
\newblock


\bibitem[Li et~al\mbox{.}(2022b)]%
        {liNeural3DVideo2022}
\bibfield{author}{\bibinfo{person}{Tianye Li}, \bibinfo{person}{Mira
  Slavcheva}, \bibinfo{person}{Michael Zollh{\"o}fer}, \bibinfo{person}{Simon
  Green}, \bibinfo{person}{Christoph Lassner}, \bibinfo{person}{Changil Kim},
  \bibinfo{person}{Tanner Schmidt}, \bibinfo{person}{Steven Lovegrove},
  \bibinfo{person}{Michael Goesele}, \bibinfo{person}{Richard Newcombe}, {and}
  \bibinfo{person}{Zhaoyang Lv}.} \bibinfo{year}{2022}\natexlab{b}.
\newblock \showarticletitle{Neural {{3D Video Synthesis From Multi-View
  Video}}}. In \bibinfo{booktitle}{\emph{Proceedings of the {{IEEE}}/{{CVF
  Conference}} on {{Computer Vision}} and {{Pattern Recognition}}}}.
  \bibinfo{pages}{5521--5531}.
\newblock


\bibitem[Li et~al\mbox{.}(2024a)]%
        {liSpacetimeGaussianFeature2024}
\bibfield{author}{\bibinfo{person}{Zhan Li}, \bibinfo{person}{Zhang Chen},
  \bibinfo{person}{Zhong Li}, {and} \bibinfo{person}{Yi Xu}.}
  \bibinfo{year}{2024}\natexlab{a}.
\newblock \showarticletitle{Spacetime {{Gaussian Feature Splatting}} for
  {{Real-Time Dynamic View Synthesis}}}. In
  \bibinfo{booktitle}{\emph{Proceedings of the {{IEEE}}/{{CVF Conference}} on
  {{Computer Vision}} and {{Pattern Recognition}}}}.
  \bibinfo{pages}{8508--8520}.
\newblock


\bibitem[Lin et~al\mbox{.}(2024)]%
        {linGaussianFlow4DReconstruction2024}
\bibfield{author}{\bibinfo{person}{Youtian Lin}, \bibinfo{person}{Zuozhuo Dai},
  \bibinfo{person}{Siyu Zhu}, {and} \bibinfo{person}{Yao Yao}.}
  \bibinfo{year}{2024}\natexlab{}.
\newblock \showarticletitle{Gaussian-{{Flow}}: {{4D Reconstruction}} with
  {{Dynamic 3D Gaussian Particle}}}. In \bibinfo{booktitle}{\emph{Proceedings
  of the {{IEEE}}/{{CVF Conference}} on {{Computer Vision}} and {{Pattern
  Recognition}}}}. \bibinfo{pages}{21136--21145}.
\newblock


\bibitem[Luiten et~al\mbox{.}(2024)]%
        {luiten2023dynamic}
\bibfield{author}{\bibinfo{person}{Jonathon Luiten}, \bibinfo{person}{Georgios
  Kopanas}, \bibinfo{person}{Bastian Leibe}, {and} \bibinfo{person}{Deva
  Ramanan}.} \bibinfo{year}{2024}\natexlab{}.
\newblock \showarticletitle{Dynamic {{3D}} Gaussians: {{Tracking}} by
  Persistent Dynamic View Synthesis}. In \bibinfo{booktitle}{\emph{{{3DV}}}}.
\newblock


\bibitem[Seidenschwarz et~al\mbox{.}(2025)]%
        {seidenschwarzDynOMoOnlinePoint2025}
\bibfield{author}{\bibinfo{person}{Jenny Seidenschwarz},
  \bibinfo{person}{Qunjie Zhou}, \bibinfo{person}{Bardienus Duisterhof},
  \bibinfo{person}{Deva Ramanan}, {and} \bibinfo{person}{Laura
  {Leal-Taix{\'e}}}.} \bibinfo{year}{2025}\natexlab{}.
\newblock \bibinfo{title}{{{DynOMo}}: {{Online Point Tracking}} by {{Dynamic
  Online Monocular Gaussian Reconstruction}}}.
\newblock
\showeprint[arxiv]{2409.02104}~[cs]
\href{https://doi.org/10.48550/arXiv.2409.02104}{doi:\nolinkurl{10.48550/arXiv.2409.02104}}


\bibitem[Stearns et~al\mbox{.}(2024)]%
        {stearnsDynamicGaussianMarbles2024}
\bibfield{author}{\bibinfo{person}{Colton Stearns}, \bibinfo{person}{Adam
  Harley}, \bibinfo{person}{Mikaela Uy}, \bibinfo{person}{Florian Dubost},
  \bibinfo{person}{Federico Tombari}, \bibinfo{person}{Gordon Wetzstein}, {and}
  \bibinfo{person}{Leonidas Guibas}.} \bibinfo{year}{2024}\natexlab{}.
\newblock \showarticletitle{Dynamic {{Gaussian Marbles}} for {{Novel View
  Synthesis}} of {{Casual Monocular Videos}}}. In
  \bibinfo{booktitle}{\emph{{{SIGGRAPH Asia}} 2024 {{Conference Papers}}}}.
\newblock


\bibitem[Sun et~al\mbox{.}(2024)]%
        {sun3DGStreamFlyTraining2024}
\bibfield{author}{\bibinfo{person}{Jiakai Sun}, \bibinfo{person}{Han Jiao},
  \bibinfo{person}{Guangyuan Li}, \bibinfo{person}{Zhanjie Zhang},
  \bibinfo{person}{Lei Zhao}, {and} \bibinfo{person}{Wei Xing}.}
  \bibinfo{year}{2024}\natexlab{}.
\newblock \showarticletitle{{{3DGStream}}: {{On-the-Fly Training}} of {{3D
  Gaussians}} for {{Efficient Streaming}} of {{Photo-Realistic Free-Viewpoint
  Videos}}}. In \bibinfo{booktitle}{\emph{Proceedings of the {{IEEE}}/{{CVF
  Conference}} on {{Computer Vision}} and {{Pattern Recognition}}}}.
  \bibinfo{pages}{20675--20685}.
\newblock


\bibitem[Wang et~al\mbox{.}(2023)]%
        {wangTrackingEverythingEverywhere2023}
\bibfield{author}{\bibinfo{person}{Qianqian Wang}, \bibinfo{person}{Yen-Yu
  Chang}, \bibinfo{person}{Ruojin Cai}, \bibinfo{person}{Zhengqi Li},
  \bibinfo{person}{Bharath Hariharan}, \bibinfo{person}{Aleksander Holynski},
  {and} \bibinfo{person}{Noah Snavely}.} \bibinfo{year}{2023}\natexlab{}.
\newblock \showarticletitle{Tracking {{Everything Everywhere All}} at
  {{Once}}}. In \bibinfo{booktitle}{\emph{Proceedings of the {{IEEE}}/{{CVF
  International Conference}} on {{Computer Vision}}}}.
  \bibinfo{pages}{19795--19806}.
\newblock


\bibitem[Wu et~al\mbox{.}(2024)]%
        {wu4DGaussianSplatting2024}
\bibfield{author}{\bibinfo{person}{Guanjun Wu}, \bibinfo{person}{Taoran Yi},
  \bibinfo{person}{Jiemin Fang}, \bibinfo{person}{Lingxi Xie},
  \bibinfo{person}{Xiaopeng Zhang}, \bibinfo{person}{Wei Wei},
  \bibinfo{person}{Wenyu Liu}, \bibinfo{person}{Qi Tian}, {and}
  \bibinfo{person}{Xinggang Wang}.} \bibinfo{year}{2024}\natexlab{}.
\newblock \showarticletitle{{{4D Gaussian Splatting}} for {{Real-Time Dynamic
  Scene Rendering}}}. In \bibinfo{booktitle}{\emph{Proceedings of the
  {{IEEE}}/{{CVF Conference}} on {{Computer Vision}} and {{Pattern
  Recognition}}}}. \bibinfo{pages}{20310--20320}.
\newblock


\bibitem[Xu et~al\mbox{.}(2024)]%
        {xuRepresentingLongVolumetric2024}
\bibfield{author}{\bibinfo{person}{Zhen Xu}, \bibinfo{person}{Yinghao Xu},
  \bibinfo{person}{Zhiyuan Yu}, \bibinfo{person}{Sida Peng},
  \bibinfo{person}{Jiaming Sun}, \bibinfo{person}{Hujun Bao}, {and}
  \bibinfo{person}{Xiaowei Zhou}.} \bibinfo{year}{2024}\natexlab{}.
\newblock \showarticletitle{Representing {{Long Volumetric Video}} with
  {{Temporal Gaussian Hierarchy}}}.
\newblock \bibinfo{journal}{\emph{ACM Trans. Graph.}} \bibinfo{volume}{43},
  \bibinfo{number}{6} (\bibinfo{year}{2024}), \bibinfo{pages}{171:1--171:18}.
\newblock
\showISSN{0730-0301}


\bibitem[Yun et~al\mbox{.}(2025)]%
        {yunCompensatingSpatiotemporallyInconsistent2025}
\bibfield{author}{\bibinfo{person}{Youngsik Yun}, \bibinfo{person}{Jeongmin
  Bae}, \bibinfo{person}{Hyunseung Son}, \bibinfo{person}{Seoha Kim},
  \bibinfo{person}{Hahyun Lee}, \bibinfo{person}{Gun Bang}, {and}
  \bibinfo{person}{Youngjung Uh}.} \bibinfo{year}{2025}\natexlab{}.
\newblock \showarticletitle{Compensating Spatiotemporally Inconsistent
  Observations for Online Dynamic {{3D}} Gaussian Splatting}. In
  \bibinfo{booktitle}{\emph{{{ACM SIGGRAPH}} 2025 Conference Papers}}
  \emph{(\bibinfo{series}{Siggraph '25})}.
\newblock


\bibitem[Zhang et~al\mbox{.}(2021)]%
        {zhangEditableFreeviewpointVideo2021}
\bibfield{author}{\bibinfo{person}{Jiakai Zhang}, \bibinfo{person}{Xinhang
  Liu}, \bibinfo{person}{Xinyi Ye}, \bibinfo{person}{Fuqiang Zhao},
  \bibinfo{person}{Yanshun Zhang}, \bibinfo{person}{Minye Wu},
  \bibinfo{person}{Yingliang Zhang}, \bibinfo{person}{Lan Xu}, {and}
  \bibinfo{person}{Jingyi Yu}.} \bibinfo{year}{2021}\natexlab{}.
\newblock \showarticletitle{Editable Free-Viewpoint Video Using a Layered
  Neural Representation}.
\newblock \bibinfo{journal}{\emph{ACM Transactions on Graphics}}
  \bibinfo{volume}{40}, \bibinfo{number}{4} (\bibinfo{year}{2021}),
  \bibinfo{pages}{149:1--149:18}.
\newblock
\showISSN{0730-0301}


\bibitem[Zhang et~al\mbox{.}(2018)]%
        {zhang2018perceptual}
\bibfield{author}{\bibinfo{person}{Richard Zhang}, \bibinfo{person}{Phillip
  Isola}, \bibinfo{person}{Alexei~A Efros}, \bibinfo{person}{Eli Shechtman},
  {and} \bibinfo{person}{Oliver Wang}.} \bibinfo{year}{2018}\natexlab{}.
\newblock \showarticletitle{The Unreasonable Effectiveness of Deep Features as
  a Perceptual Metric}. In \bibinfo{booktitle}{\emph{CVPR}}.
\newblock


\bibitem[Zhou et~al\mbox{.}(2024)]%
        {zhouDrivingGaussianCompositeGaussian2024}
\bibfield{author}{\bibinfo{person}{Xiaoyu Zhou}, \bibinfo{person}{Zhiwei Lin},
  \bibinfo{person}{Xiaojun Shan}, \bibinfo{person}{Yongtao Wang},
  \bibinfo{person}{Deqing Sun}, {and} \bibinfo{person}{Ming-Hsuan Yang}.}
  \bibinfo{year}{2024}\natexlab{}.
\newblock \showarticletitle{{{DrivingGaussian}}: {{Composite Gaussian
  Splatting}} for {{Surrounding Dynamic Autonomous Driving Scenes}}}. In
  \bibinfo{booktitle}{\emph{Proceedings of the {{IEEE}}/{{CVF Conference}} on
  {{Computer Vision}} and {{Pattern Recognition}}}}.
  \bibinfo{pages}{21634--21643}.
\newblock


\end{thebibliography}

\end{document}